\documentclass[10pt,twocolumn,a4paper]{article}
\usepackage[utf8]{inputenc}
\usepackage[T1]{fontenc}
\usepackage[american]{babel}

\usepackage{textcomp}
\usepackage[activate={true,nocompatibility},final,tracking=false,kerning=true,spacing=false,factor=1100,stretch=10,shrink=10]{microtype}
\usepackage{etoolbox}
\apptocmd{\sloppy}{\hbadness 10000\relax}{}{}

\usepackage{cite}
\usepackage{url}
\usepackage{graphicx}
\usepackage{authblk}

\usepackage{algorithm}
\usepackage{algorithmic}

\usepackage{booktabs}

\usepackage{siunitx}
\usepackage{physics}

\usepackage{hyperref}

\usepackage{stfloats}

\title{Towards detection and classification of microscopic foraminifera using transfer learning}
\author[1]{Thomas Haugland Johansen\thanks{Corresponding Author: thomas.h.johansen@uit.no}}
\author[2]{Steffen Aagaard Sørensen}
\affil[1]{Department of Mathematics and Statistics, UiT The Arctic University of Norway}
\affil[2]{Department of Geology, UiT The Arctic University of Norway}

\date{\vspace{-5ex}}

\begin{document}
\maketitle

\begin{abstract}
Foraminifera are single-celled marine organisms, which may have a planktic or benthic lifestyle.
During their life cycle they construct shells consisting of one or more chambers, and these shells remain as fossils in marine sediments.
Classifying and counting these fossils have become an important tool in e.g. oceanography and climatology.
Currently the process of identifying and counting microfossils is performed manually using a microscope and is very time consuming.
Developing methods to automate this process is therefore considered important across a range of research fields.
The first steps towards developing a deep learning model that can detect and classify microscopic foraminifera are proposed.
The proposed model is based on a VGG16 model that has been pretrained on the ImageNet dataset, and adapted to the foraminifera task using transfer learning.
Additionally, a novel image dataset consisting of microscopic foraminifera and sediments from the Barents Sea region is introduced.
\end{abstract}

\section{Introduction}

Foraminifera are ubiquitous ocean dwelling single-celled microorganisms that may have a planktic (living in the water column) or benthic (living at or within the seabed) lifestyle.
During their life cycle foraminifera construct shells with one or more chambers.
The shells are commonly composed of calcium carbonate (calcareous foraminifera) or constructed from sediment particles cemented together (agglutinated foraminifera).
They are recognizable due to their interspecies morphological differences.
The shells remain in the marine sediments as fossils, and can be extracted from rock or marine sediment samples.
Foraminifera are common in both modern and ancient environments and have become invaluable tools in oceanographic and geoscience research as well as in petroleum exploration.
For example in paleo-research, fossilized foraminiferal fauna compositions and/or chemical composition of individual shells are frequently used to infer past changes in ocean temperature, salinity, ocean chemistry, and global ice volume~\cite{hald2007variations, spielhagen2011enhanced, aagaard2014late}.
In ecotoxicology and pollution monitoring studies, changes in foraminiferal abundance, morphology and faunal composition are used for detecting ecosystem contamination~\cite{frontalini2011benthic}.
In the petroleum industry, foraminiferal analysis is an important tool to infer ages and paleoenvironments of sedimentary strata in oil wells during exploration, which aids the detection of potential hydrocarbon deposits~\cite{boardman1987fossil, singh2008micropaleontology}.

Statistical counting of foraminifera species, their number and distribution, represents important data for marine geological climate and environmental research and in petroleum exploration.
Counting, identification and picking of foraminifera in prepared sediment samples using a microscope is a very time and resource demanding process, which has practically been conducted the same way since the use of microscope foraminiferal studies started in the early 1800's.
Progress in deep learning makes it possible to automate this work, which will contribute to better quality, higher quantity, reduced resource usage, and more cost effective data collection.
Existing research groups have already started with image recognition of foraminifera~\cite{mitra2019automated, zhong2017comparative, ge2017coarse, de2017automatic}, but the training data currently needs to be ``tailor made'' with the most abundant foraminiferal species for a specific geographical region.

\section{Transfer learning}

There are a number of transfer learning methods used in deep learning, and in the presented experiments two such methods are implemented, namely feature extraction and fine tuning.

The strengths of a deep convolutional neural network~(CNN) model is its many layers of filters, learned by training on millions of images~\cite{simonyan2014deep}.
Learning the weights of these layers can require an enormous amount of images, depending on e.g. the depth and complexity the model, the input domain, etc.
However, the learned filters represent somewhat abstract feature detectors that can be transferred to new domains~\cite{bengio2013representation,yosinski2014transferable}.
In other words, it is possible to re-use the weights of a pretrained CNN model for new classification tasks.
In its simplest form this is achieved by using the convolutional blocks of the model as a feature extractor, and the extracted features can then be passed to any classifier.
The weights of the classifier need to be learned, but the weights of the pretrained filter layers are preserved or ``frozen''.
Typically the classifier is chosen such that it performs well at the task of predicting output labels using the extracted features, while also being tractable to train.

It is also possible to re-train some layers of the CNN to optimize the extracted features to the new domain, which is referred to as fine tuning.
This will then be a trade-off between adapting the pretrained model to the new image modalities, but with the risk of overfitting given the typically small size of the training dataset.
Which layers to re-train typically depend on several factors, such as similarity between the new and the original image modalities.

\section{Monte Carlo dropout}

The complexity of a CNN classifier makes the output inconceivable in terms of the usual image feature interpretation, and there is a need for a measure of uncertainty.
A step in that direction is to allow for stochastic prediction through Monte Carlo dropout.

\begin{figure*}[t]
        \centering
        \includegraphics[width=0.7\linewidth,trim={0 1cm 0 1cm},clip]{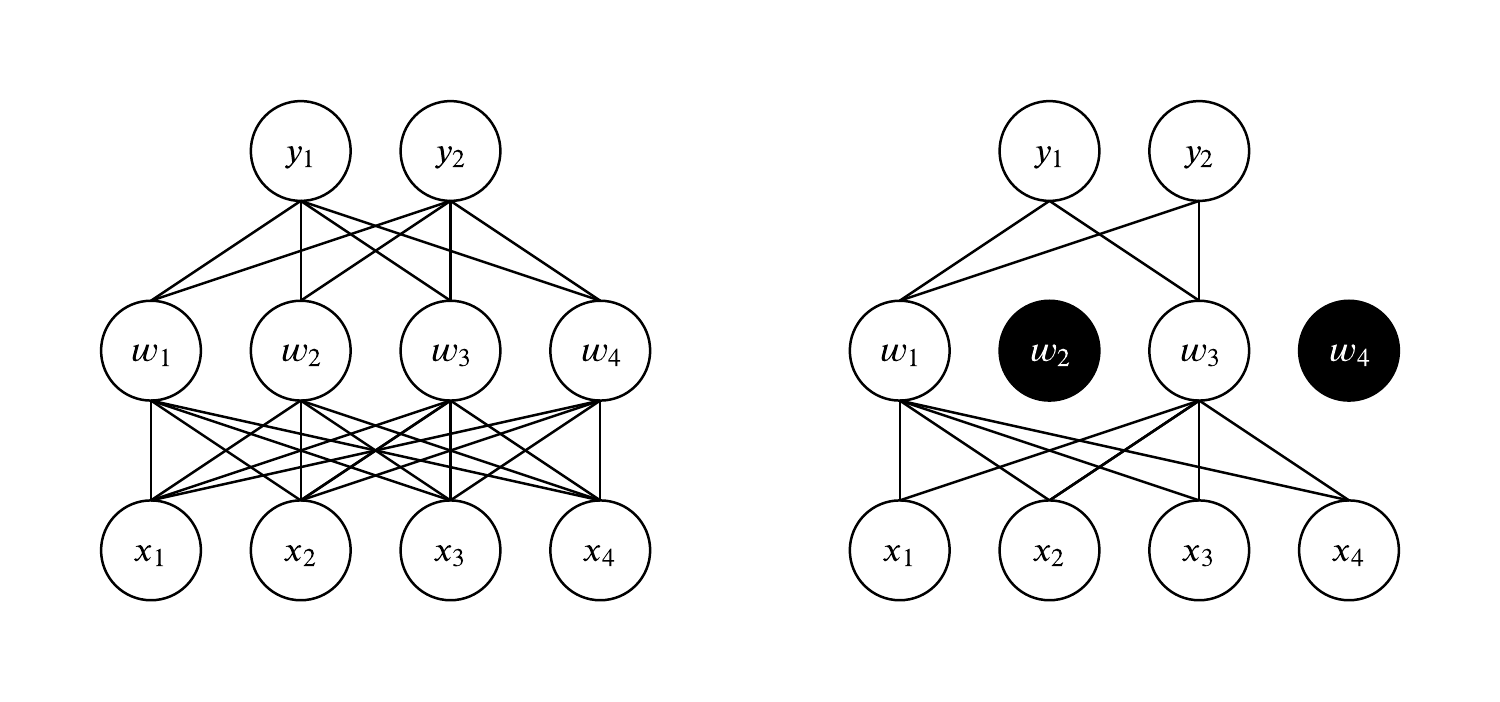}
        \caption{Toy example illustrating a neural network with and without dropout applied.}
        \label{fig:dropout}
    \end{figure*}

Dropout is a regularization technique frequently used when training deep neural network models to reduce the chance of overfitting~\cite{srivastava2014dropout}.
The basic idea is that a specified percentage of weights for some layers in the model are set to zero, effectively turning off the corresponding units in that layer.
This percentage is referred to as the dropout rate and is considered a model hyperparameter.
Which units to drop during training are chosen at random, typically sampling from a uniform distribution.
One intuition behind dropout is that it helps prevent units from co-adapting, which might otherwise lead to ``memorization'' of training data.
See Figure~\ref{fig:dropout} for an illustrative toy example of how dropout behaves with a rate of $50\%$.

Once the model has been trained, the dropout rate is normally set to zero to ensure predictions are deterministic.
Since units are dropped at random, predictions are stochastic, and this is the underlying idea of Monte Carlo dropout~\cite{gal2016dropout}.
By considering dropout to be a Bayesian approximator in some sense, it becomes possible to analyze e.g. model uncertainty.

Assume a neural network $f$ with model parameters $\vb{W}$ has been trained such that
    \begin{gather}
        \vb{\tilde{Y}} = f\qty(\vb{X}; \vb{W}) ,
    \end{gather}
where $\vb{\tilde{Y}}$ is the predicted output for some dataset $\vb{X}$ with true output $\vb{Y}$.
Monte Carlo dropout can then be implemented by iterating over the dataset $N$ times collecting the output predictions,
    \begin{gather}
        \vb{\tilde{Y}}_i = f\qty(\vb{X}; \vb{W}_i) , \quad i=1, \dotsc, N \label{eqn:preds}
    \end{gather}
where $\vb{W}_i$ represents the model parameters for the $i$-th iteration after applying dropout.
Using the collected predictions, Monte Carlo estimates of the predictive mean and variance can be computed,
    \begin{gather}
        \vb*{\tilde{\mu}} = \frac{1}{N} \sum_{i=1}^N \vb{\tilde{Y}}_i , \label{eqn:pred-mean} \\
        \vb*{\tilde{\sigma}} = \frac{1}{N} \sum_{i=1}^N \qty( \vb{\tilde{Y}}_i - \vb*{\tilde{\mu}} )^2 . \label{eqn:pred-var}
    \end{gather}
The predictive mean $\vb*{\tilde{\mu}}$ can be interpreted as the ensemble prediction for $N$ different models.
Similarly, the uncertainty of the ensemble predictions can be expressed using the predictive variance.

\section{Preparing the datasets}

The materials (foraminifera and sediment) used for the present study were collected from sediment cores retrieved in the Arctic Barents Sea region.
In order to achieve a good representation of the planktic and benthic foraminiferal fauna of the area, the specimens were picked from sediments influenced by Atlantic, Arctic, polar, and coastal waters representing different ecological environments.
Foraminiferal specimens (planktics, benthics, agglutinated benthics) were picked from the \SIrange{100}{1000}{\micro\metre} size fraction of freeze dried and subsequently wet sieved sediments.
Sediment grains representing a common sediment matrix were also sampled from the \SIrange{100}{1000}{\micro\metre} size range.
The basis for the datasets were collected by photographing either pure benthic (calcareous or agglutinated), planktic assemblages, or sediments containing no foraminiferal specimens.
In other words, each image contained only specimens belonging to one of four high-level classes; planktic, calcareous benthic, agglutinated benthic, sediment.
This approach simplified the task of labeling each individual specimen with the correct class.
All images were captured with a 5 megapixel Leica DFC450 digital camera mounted on a Leica microscope.

\begin{figure*}[ht]
        \centering
        \includegraphics[width=0.7\linewidth]{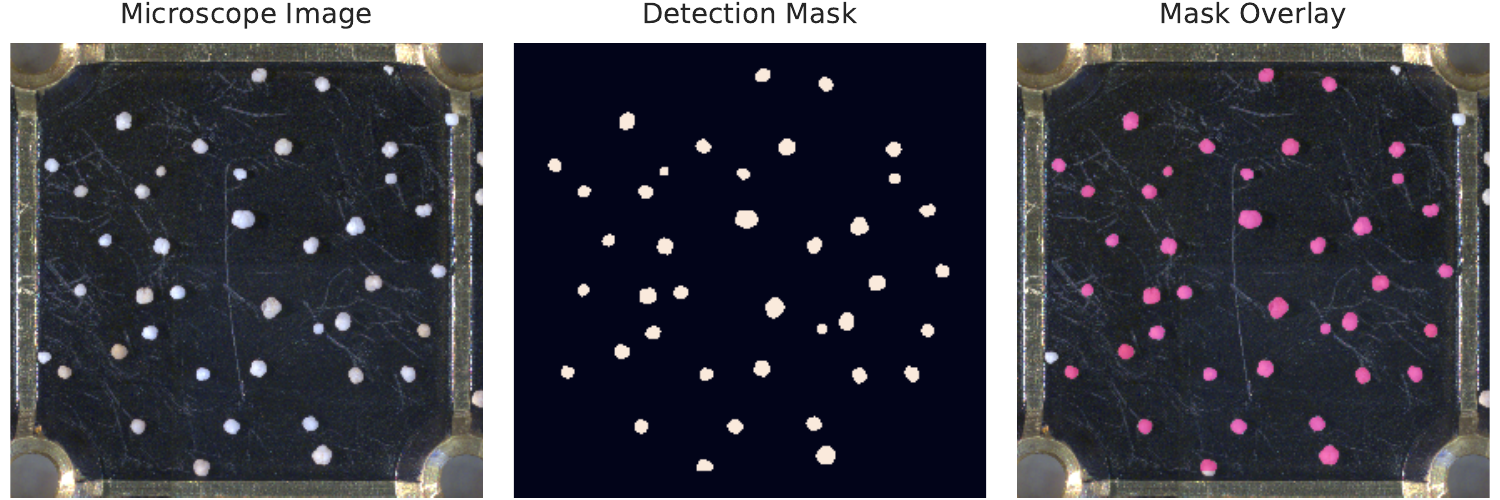}
        \caption{Examples from the detection and extraction procedure used to create the foraminifera dataset.}
        \label{fig:pipeline}
    \end{figure*}

From each of the images collected from the microscope, smaller images of each individual specimen were extracted using a very simple, yet effective, object detection scheme based on Gaussian filtering, grayscale thresholding, binary masking and connected components.
The first pass of Gaussian filtering, grayscale thresholding and binary masking was tuned to remove the metallic border present in each image, which can be seen in Figure~\ref{fig:pipeline}.
The next pass of filtering, thresholding and masking was tuned to detect the foraminifera and sediment candidates.
Very small objects, which included remnant particulates (considered noise) from e.g. damaged specimens, were discarded based on the number of connected components; all candidates with less than 1024 pixels were discarded.
After selecting candidates from the original microscope images, all of the individual specimen images were extracted by placing a $224 \!\times\! 224$ pixel crop region at the ``center of mass'' of each candidate.
An example from this process can be seen in Figure~\ref{fig:pipeline}.

\begin{figure*}[b]
        \centering
        \includegraphics[width=0.8\linewidth,trim={0 0.4cm 0 0},clip]{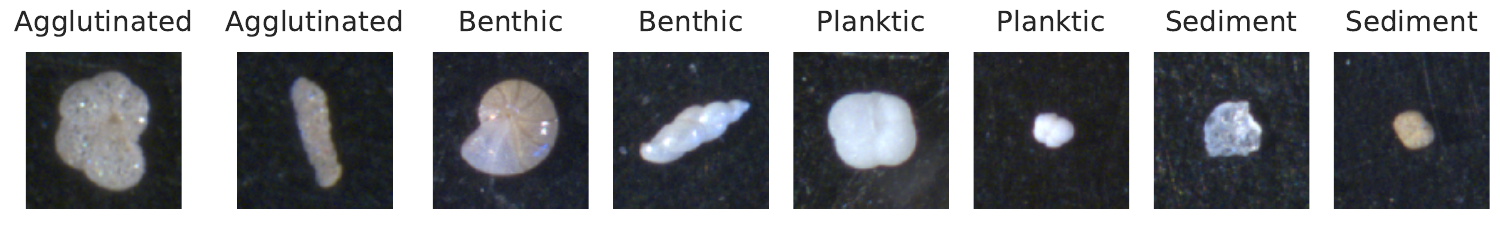}
        \caption{Examples of typical specimens from each of the four categories found in the image dataset.}
        \label{fig:examples}
    \end{figure*}

Upon completing the object detection and image extraction procedure, the result was a dataset containing a total of 2673 images.
These images were then stratified into training, validation and test sets using a 80/10/10 split.
Examples of extracted images can be seen in Figure~\ref{fig:examples}.

\section{Experiments}

All experiments presented are based on a VGG16~\cite{simonyan2014deep} model that had been pretrained on the ImageNet~\cite{deng2009imagenet} dataset.
The choice of model was made primarily due to prior experience and familiarity with the architecture.

\subsection{Model design and training}

\begin{figure*}[t]
        \centering
        \includegraphics[width=\linewidth]{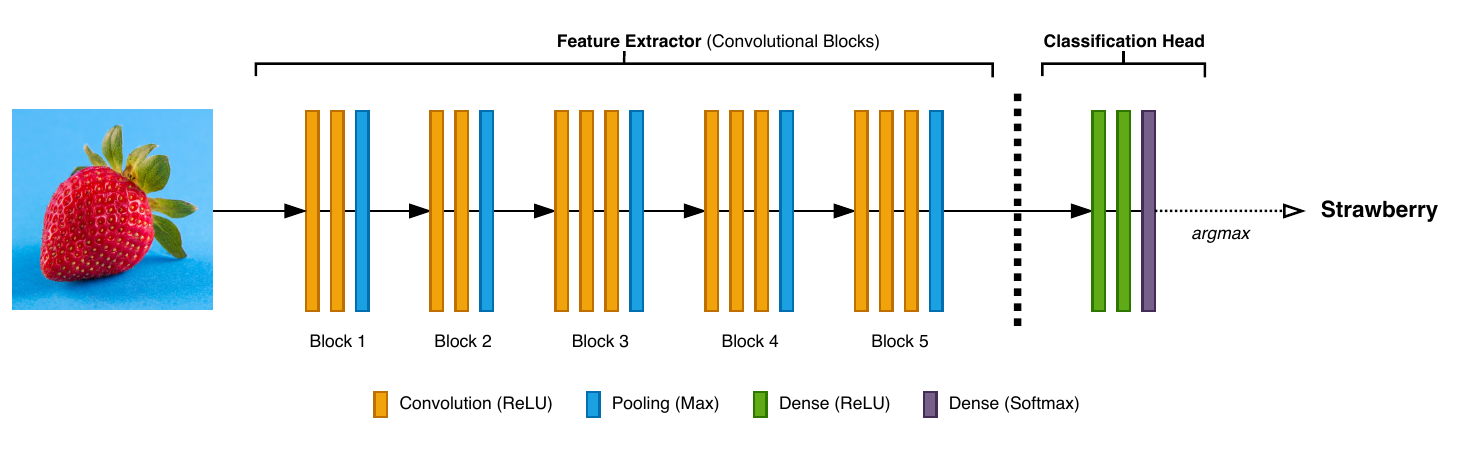}
        \caption{Simplified architecture diagram of the VGG16 model. Input images are passed through the convolutional blocks, and feature vectors are then transformed by dense layers into softmax predictions.}
        \label{fig:vgg16}
    \end{figure*}

Using a pretrained VGG16 model, feature vectors were extracted from each of the foraminifera and sediment images in the dataset.
See Figure~\ref{fig:vgg16} for a simplified illustration of the VGG16 model architecture.
The feature extraction procedure was done by removing the fully-connected dense layers, the so called ``classification head'', at the end of the VGG16 model.
Feature vectors were then extracted from the last convolutional block, and used as input features to a new deep neural network model designed to classify foraminifera and sediment.
This new classification model went through several designs during initial prototyping, varying in number of layers and units per layer.
Ultimately, hyperparameter tuning was performed to finalize the design of the classifier.
This was done using a grid search approach, which tested 72 different permutations of units per layer, dropout rate, and optimization algorithm.
The final end-to-end model architecture can be seen summarized in Table~\ref{tab:model}.
    \begin{table}
        \centering
        {\small\begin{tabular}{lcc}
            \toprule
            Layer Type & Input Dim. & Output Dim. \\
            \midrule
            VGG16   & $224 \!\times\! 224 \!\times\! 3$ & $7 \!\times\! 7 \!\times\! 512$ \\
            \midrule
            Dense (ReLU)     & $25088$ & $512$ \\
            Dense (ReLU)     & $512$   & $64$ \\
            Dense (Softmax)  & $64$    & $4$
        \end{tabular}}
        \caption{High-level summary of the deep learning model used to classify foraminifera and sediments.}
        \label{tab:model}
    \end{table}

The model was first trained with all weights for the VGG16 model being fixed, and thus only the weights of the new classification head were optimized.
All training was done using a batch size of 32, cross entropy loss, and an \emph{Adam}~\cite{kingma2014adam} optimizer with an initial learning rate of $10^{-4}$.
To reduce the chance of overfitting, early stopping was implemented based on the validation accuracy computed at the end of each training epoch.
On average, due to early stopping, each training session stopped after 7 epochs, with each epoch consisting of 260 training steps.
After initial training of the classification model on feature vectors extracted from the VGG16 model, fine-tuning was implemented to improve classification accuracy.
This was achieved by ``unfreezing'' the last two convolutional blocks of the VGG16 model, thus allowing the model to specialize those parameters to the new classification task.
The initial learning rate during fine-tuning was reduced to $10^{-7}$ to ensure smaller, incremental gradient updates.

Given the relatively small dataset, image augmentation was implemented to synthetically boost the number of training images.
The augmentations consisted of flipping, rotating, as well as changing brightness, contrast, hue, and saturation.
Flipping was done horizontally, and rotations in increments of 90 degrees.
Brightness, contrast and saturation values were randomly augmented by $\pm10\%$, whereas hue was augmented by $\pm5\%$.
These augmentations were chosen based on qualitative analysis of the dataset to ensure they were both representative and valid.
Each augmentation was applied in a randomized fashion to every image in a batch, each time a training batch was sampled.

The training procedure was repeated multiple times to reduce the effects of random initialization of model weights.
After only training the classification head, the mean accuracy on the test data was $97.0 \pm 0.6\%$.
Fine-tuning improved the results to a mean accuracy of $98.8 \pm 0.2\%$.

\subsection{Model analysis}

After training, Monte Carlo dropout was implemented in order to investigate and analyze the trained models.
Model predictions were collected as expressed in \eqref{eqn:preds} for $N=100$, with all dropout layers turned on and using the entire test set.
Predictive mean and variance were calculated using \eqref{eqn:pred-mean} and \eqref{eqn:pred-var}, respectively.

\begin{figure}[b]
        \centering
        \includegraphics[width=\linewidth]{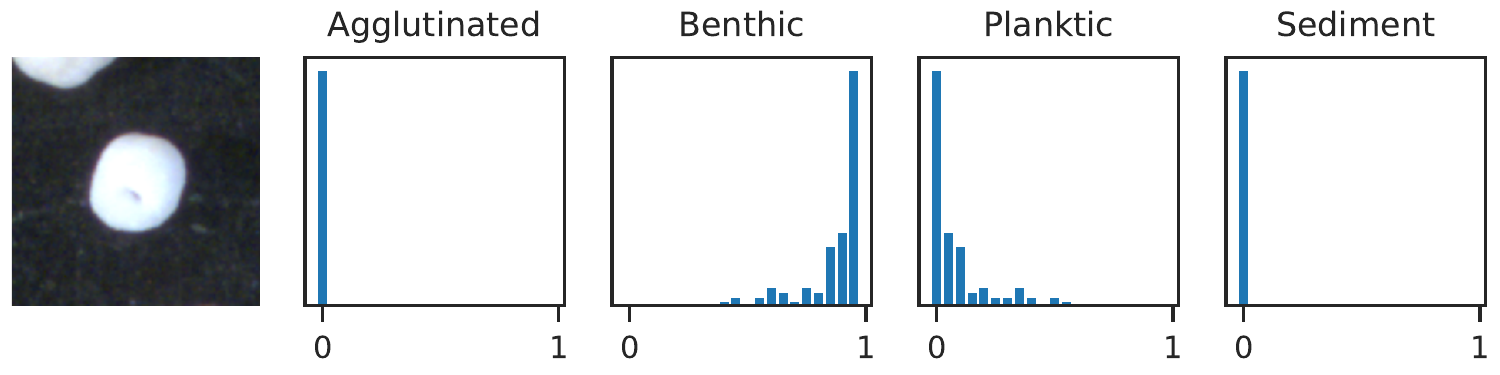}
        \caption{Overexposed planktic foraminifera, misclassified as benthic. Histograms represent distributions of softmax predictions from MC dropout.}
        \label{fig:specimen}
    \end{figure}

Using these results made it possible to uncover difficult cases in the dataset where the model was having problems with the classification.
There were two scenarios; the model was uncertain about the prediction, or it was certain, but the prediction was incorrect.
When studied qualitatively, some of the challenging images contained overexposed specimens that were missing details such as texture.
In other cases, specimens were oriented in such a way that the morphological characteristics of the foraminifera were not visible.
An example of an overexposed specimen can be seen in Figure~\ref{fig:specimen}.
Some of the challenging cases were shown to a trained expert, which was able to correctly classify all specimens.

The mean accuracy for all Monte Carlo simulations was $97.9 \pm 0.5\%$.
Furthermore, by considering each simulation to be part of an ensemble of models with a majority voting scheme, the accuracy of the ensemble predictions was $98.5\%$.
These results are comparable to the model without Monte Carlo dropout.

\section{Concluding remarks}



Based on the presented experiments it is clear that training deep learning models to accurately classify microscopic foraminifera is possible.
Using VGG16 pretrained on ImageNet to extract features from foraminifera produces very promising results, which can then be further improved by fine-tuning the pretrained model.
The results are comparable to equivalent efforts by other research group using different datasets of foraminifera and sediments.


To uncover images in the dataset that the model is uncertain about techniques such as Monte Carlo dropout can used.
These results can then be used to identify classes that need more training data, or perhaps alludes to further image augmentation, etc.

Future work should involve investigations using model architectures other than VGG16 should be conducted, comparing differences in prediction accuracy, computational efficiency during training and inference, and so forth.
Once bigger datasets become available, efforts should also invested towards training novel models from scratch, and comparing to pretrained models.


\bibliographystyle{abbrv}
\bibliography{refs}

\end{document}